\pdfoutput=1
\documentclass[11pt]{article}

\usepackage{EMNLP2022}

\usepackage{times}
\usepackage{latexsym}
\usepackage{amsmath}
\usepackage{multirow}
\usepackage{tcolorbox}
\usepackage{diagbox}
\usepackage[normalem]{ulem}
\useunder{\uline}{\ul}{}
\usepackage{multirow}

\usepackage[T1]{fontenc}
\usepackage[utf8]{inputenc}

\usepackage{microtype}

\usepackage{inconsolata}

\usepackage{graphicx}

\title{AstBERT: Enabling Language Model for Financial Code Understanding with Abstract Syntax Trees}
\author{\normalsize{Rong Liang} \\ \normalsize{Ant Group} \\ \normalsize{liangrong.liang@antgroup.com}
        \And
        \normalsize{Tiehua Zhang}\thanks{\hspace{0.15cm}Corresponding Author} \\ \normalsize{Ant Group} \\ \normalsize{zhangtiehua.zth@antgroup.com} \And
        \normalsize{Yujie Lu} \\ \normalsize{Ant Group} \\ \normalsize{lyj272836@antgroup.com} \AND
        \normalsize{Yuze Liu} \\ \normalsize{Ant Group} \\ \normalsize{liuyuze.liuyuze@antgroup.com} \And
        \normalsize{Zhen Huang} \\ \normalsize{Ant Group} \\ \normalsize{hz101346@antgroup.com} \And
        \normalsize{Xin Chen} \\ \normalsize{Ant Group} \\ \normalsize{jinming.cx@antgroup.com}}

\begin{document}
\maketitle
\begin{abstract}
Using the pre-trained language models to understand source codes has attracted increasing attention from financial institutions owing to the great potential to uncover financial risks. However, there are several challenges in applying these language models to solve programming language related problems directly. For instance, the shift of domain knowledge between natural language (NL) and programming language (PL) requires understanding the semantic and syntactic information from the data from different perspectives. To this end, we propose the AstBERT model, a pre-trained PL model aiming to better understand the financial codes using the abstract syntax tree (AST). Specifically, we collect a sheer number of source codes (both Java and Python) from the Alipay code repository and incorporate both syntactic and semantic code knowledge into our model through the help of code parsers, in which AST information of the source codes can be interpreted and integrated. We evaluate the performance of the proposed model on three tasks, including code question answering, code clone detection and code refinement. Experiment results show that our AstBERT achieves promising performance on three different downstream tasks.
\end{abstract}

\section{Introduction}
Programming language and source code analysis using deep learning methods have received increasing attention in recent years. Using pre-trained model such as such as BERT \cite{devlin2018bert}, AlBERT~\cite{lan2019albert} receive a great success on different NLP tasks.
%These pre-trained language models are trained on large unlabeled data by self-supervised task to effectively learn contextual representations. The success of the pre-trained model in NLP is reproduced in other related fields, such as VideoBERT \cite{sun2019videobert} in video-language and VisualBERT \cite{li2019visualbert} in picture-language. 
Inspired by that, some researchers attempt to apply this technique to comprehend source codes. For instance, CodeBERT \cite{feng2020codebert} is a pre-trained model using six different programming languages from GitHub, demonstrating a good performance comparing with different embedding techniques.

Although pre-trained models are now widely used for different purposes, it is rare to see how to apply such techniques to financial service codes. It is believed that re-training the model using financial service code could help uncover the code hazards before being released and circumvent any economic damage~\cite{guo2020graphcodebert}. Existing research points out that the use of domain knowledge is critical when it comes to training a well-performing model, and one way to solve this problem is to pre-train a model using specific domain corpora from scratch~\cite{hellendoorn2019global}. However, pre-training a model is generally time-consuming and computationally expensive, and domain corpora are often not enough for pre-training tasks, especially in the financial industry, where the number of open-sourced codes is limited.
%It is believed that source code understanding is more difficult compared to normal natural language tasks, owing to the rigorous programming syntax. Therefore, it is rational to believe that bringing structure information of source code into the model will improve the performance of code understanding.
%However, pre-training a model is generally time-consuming and computationally expensive, and domain corpora is often not enough for pre-training tasks. To address this problem, K-BERT proposed by is a knowledge-enabled language representation model with knowledge graphs (KGs), which use triples (entity and its relation) in KGs to bring domain knowledge for the origin sentence. 

To this end, we propose an AstBERT model, a pre-trained language model aiming to better understand the financial codes using abstract syntax trees (AST). To be more specific, AST is a tree structure description of code semantics. Instead of using source code directly, we leverage AST as the prominent input information when training and tuning AstBERT. To overcome the token explosion problem that usually happens when generating the AST from the large-scale code base, a pruning method is applied beforehand, followed by a designated AST-Embedding Layer to encode the pruned code syntax information. To save the training time and resources, we adopt the pre-trained CodeBERT \cite{feng2020codebert} as our inception model and continue to train on the large quantity of AST corpus. In this way, AstBERT can capture semantic information for both nature language (NL) and programming language (PL). 

We train AstBERT on both Python and Java corpus collected from Alipay code repositories, which contains about 0.2 million functions in java and 0.1 million functions in python. Then we evaluate its performance on different downstream tasks. The main contributions of this work can be summarized as follows:
% outperforms Bert, CRF baselines on code information extraction task by 3.5\%, and RoBERTa, CodeBERT baselines on code search by 4.25\%. 

\begin{itemize}	
\item We propose a simple yet effective way to enhance the pre-trained language model's ability to understand programming languages in the financial domain with the help of abstract syntax tree information.
\item We conduct extensive experiments to verify the performance of AstBERT on code-related tasks, including code question answering, code clone detection and code refinement. Experiments results show that AstBERT demonstrates a promising performance for all three downstream tasks.
% Empirical results show that AstBERT is effective on \textcolor{red}{three tasks, including code question answering, clone detection and code refinement}.
% search and code information extraction tasks.
% \item \textcolor{red}{We created a dataset for the name entity recognition(NER)} task in code, which, to our knowledge, is the first NER dataset in code.
\end{itemize}

\section{Related Work}
In this part, we describe  existing pre-trained models and datasets in code language interpretation in detail.
\subsection{Datasets in Code Understanding}
It is inevitable to leverage a high-quality dataset in order to pre-train a model that excels in code understanding. Some researchers have started to build up the dataset needed for the code search task in~\cite{nie2016query}, in which different questions and answers are collected from Stack Overflow. Also, a large-scale unlabeled text-code pairs are extracted and formed from GitHub by \cite{husain2019codesearchnet}. Three benchmark datasets are builed by \cite{heyman2020neural}, each of which consists of a code snippet collection and a set of queries. An evaluation dataset developed by \cite{li2019neural} consists of natural language question and code snippet pairs. They manually check whether the questions meet the requirements and filter out the ambiguous pairs.  A model trained by \cite{yin2018learning} on a human-annotated dataset is used to automatically mine massive natural language and code pairs from Stack Overflow. Recently, CoSQA dataset constructed by \cite{huang2021cosqa} that includes 20,604 labels for pairs of natural language queries and codes. CoSQA is annotated by human annotators and it is obtained from real-world queries and Python functions. It is rare to find open-sourced public source code in the financial domain, and we therefore retrieve both Python and Java code from the Alipay code repositories.
% From the above we can know, most of the datasets in language-code are about code search task which belongs to sequence classification task. We propose a new dataset for named entity reorganization(NER) task in code.

\subsection{Models in Code Understanding}

Using deep learning network to solve language-code tasks has been studied for years. A Multi-Modal Attention Network trained by \cite{wan2019multi} represents unstructured and structured features of source code with two LSTM. A masked language model\cite{kanade2019pre} is trained on massive Python code obtained from GitHub and used to obtain a high-quality embedding for source code. A set of embeddings \cite{karampatsis2020scelmo} based on ELMo \cite{peters2018deep} and conduct bug detection task. The results prove that even a low-dimensional embedding trained on a small corpus of programs is very useful for downstream task. Svyatkovskiy et al. use GPT-2 framework and train it from scratch on source code data to support code generative task like code completion \cite{svyatkovskiy2020intellicode}. CodeBERT \cite{feng2020codebert} is a multi-PL (programming language) pretrained model for code and natural language, and it is trained with the new learning objective based on replaced token detection. C-BERT proposed by \cite{buratti2020exploring} is pre-trained from C language source code collected from GitHub to do AST node tagging task. 

Different with previous work, AstBERT is a simple yet effective way to use pre-trained model in code interpretation field. Instead of training a large-scale model from scratch, it incorporates AST information into a common language model, from which the code understanding can be derived.

\section{AstBERT}

In this part, we describe the details about AstBERT. 
% including model architecture, including input/out layer, different embedding layers, and how to use AstBERT for downstream tasks.

\subsection{Model Architecture}

\begin{figure}
\centering
        \includegraphics[scale=0.34]{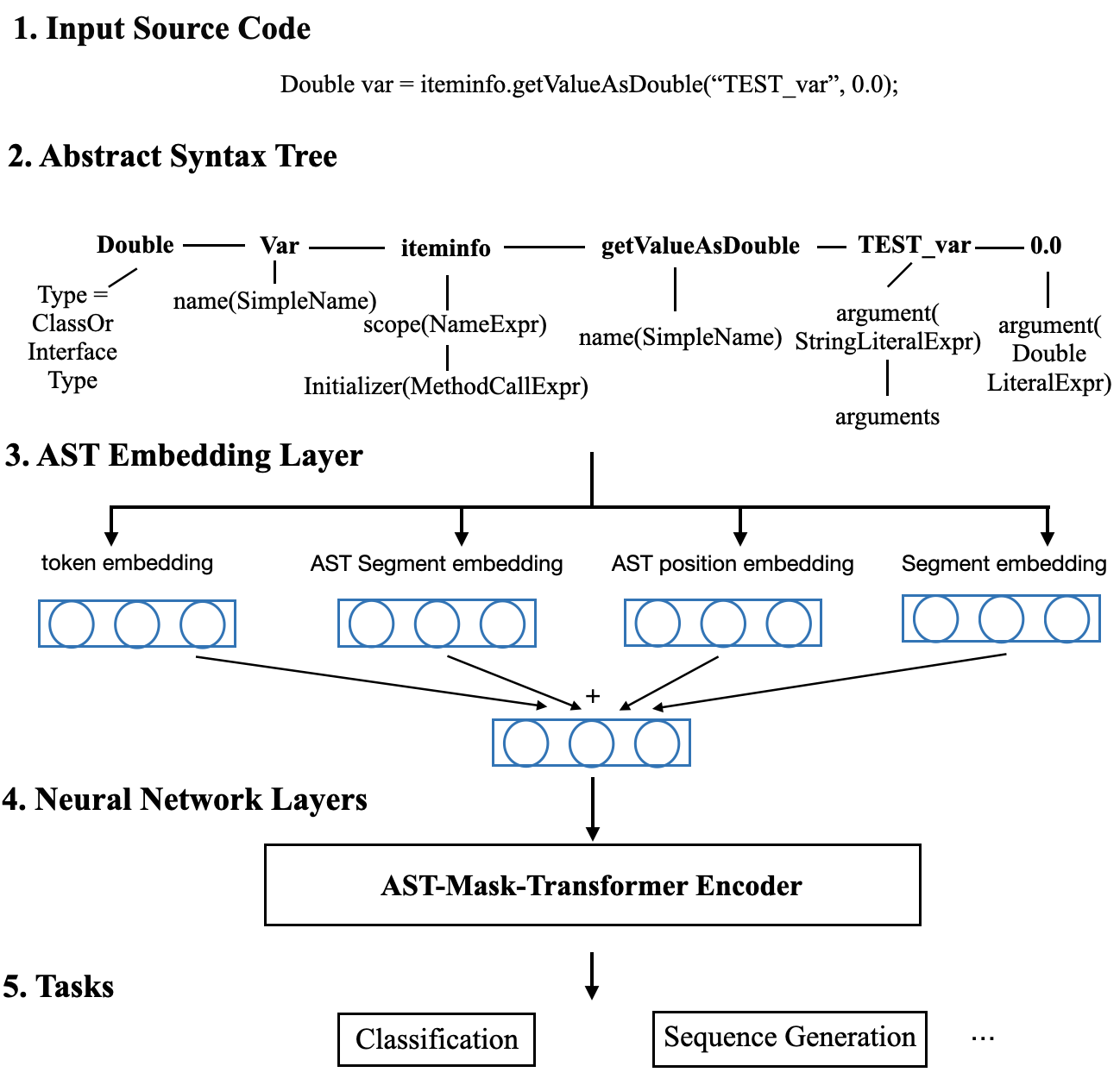}
		\caption{The model structure of AstBERT: An easy and effective way to enhance pre-trained language model's ability for code understanding}
		\vspace{-5mm}
\end{figure}

Figure 1 shows the main architecture of AstBERT. Instead of using source code directly, the pruned AST information is used as the input. For each source code token, the AST information is attached in the front, and the position index is used to show the order of the input. There are four embedding modules at the AST embedding layer. Token embedding is similar to what is in BERT \cite{devlin2018bert}, one key difference is that the vocabularies used are AST keywords. The token is then encoded to a vector format. Additionally, in AstBERT, AST-segment and AST-position are used to integrate the structure information of AST,
% tell which token is AST information and which token is source code,} 
the detail of their function will be introduced in subsection 3.3. After the AST embedding layer, the embedding vectors are then forwarded to a multi-layer bidirectional AST-Mask-Transformer encoder~\cite{vaswani2017attention} to generate hidden vectors. The difference is that we use AST-Mask-Self-Attention instead of Self-Attention to calculate the attention score, the detail of which will be unveiled in subsection 3.4.
% After being encoded by AST-Mask-Transformer encoder, 
% \textcolor{red}{we use AST-Selection layer to mask the hidden vector from tokens of AST information and retain the hidden vector from tokens of original source code.} 
In the output layer, the hidden vectors generated by AST-Mask-Transformer encoder will be used for classification or sequence generation tasks.

\subsection{Input and Pruning}

\begin{figure}[t!]
\vspace{5mm}
        \includegraphics[scale=0.44]{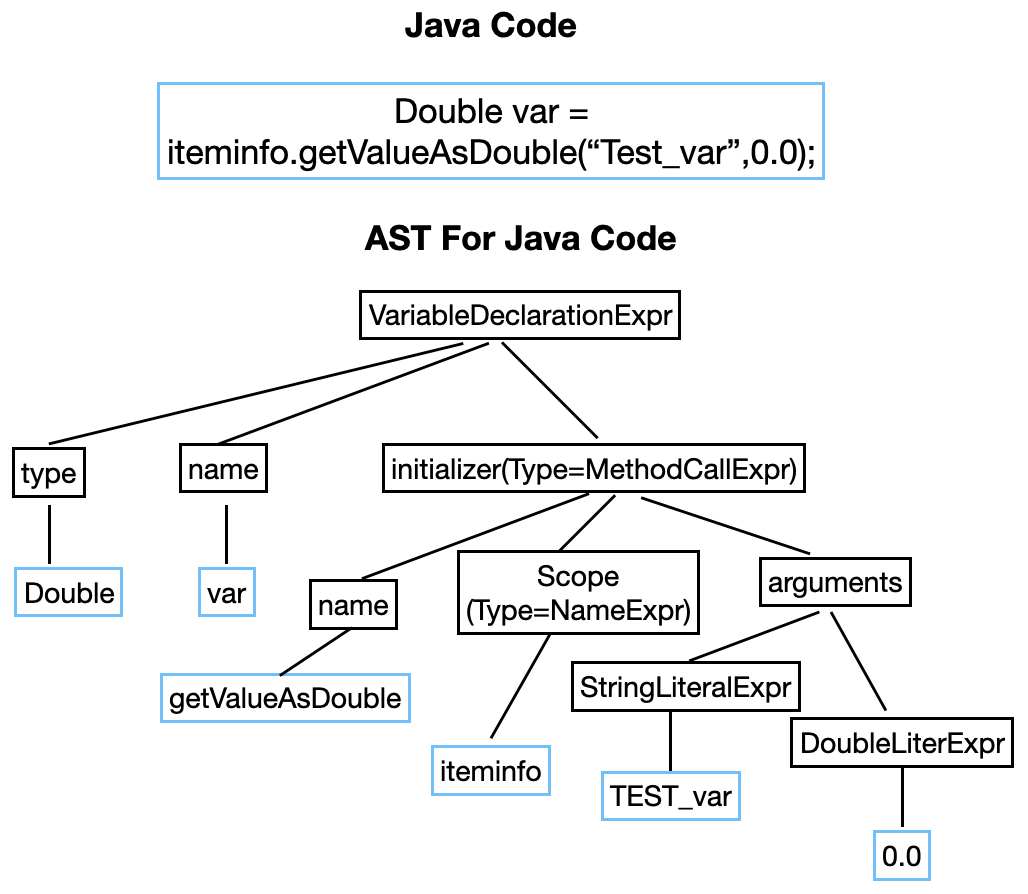}
		\caption{AST-based code representation of a financial code snippet}
\vspace{-2mm}
\end{figure}

We introduce the pruning process in this part. As shown in Figure 2, the AST contains the complete information of the source code and provide the brief description for each token. For example, the \textit{getValueAsDouble} is the name for \textit{MethodCallExpr} (one of the AST node types) and the \textit{TEST\_var} is an argument for \textit{MethodCallExpr}. We know the \textit{Double} is a type of the variable \textit{var} from AST. Such AST information reveals the semantic knowledge of the source code. 

\begin{figure}[t!]
        \includegraphics[scale=0.32]{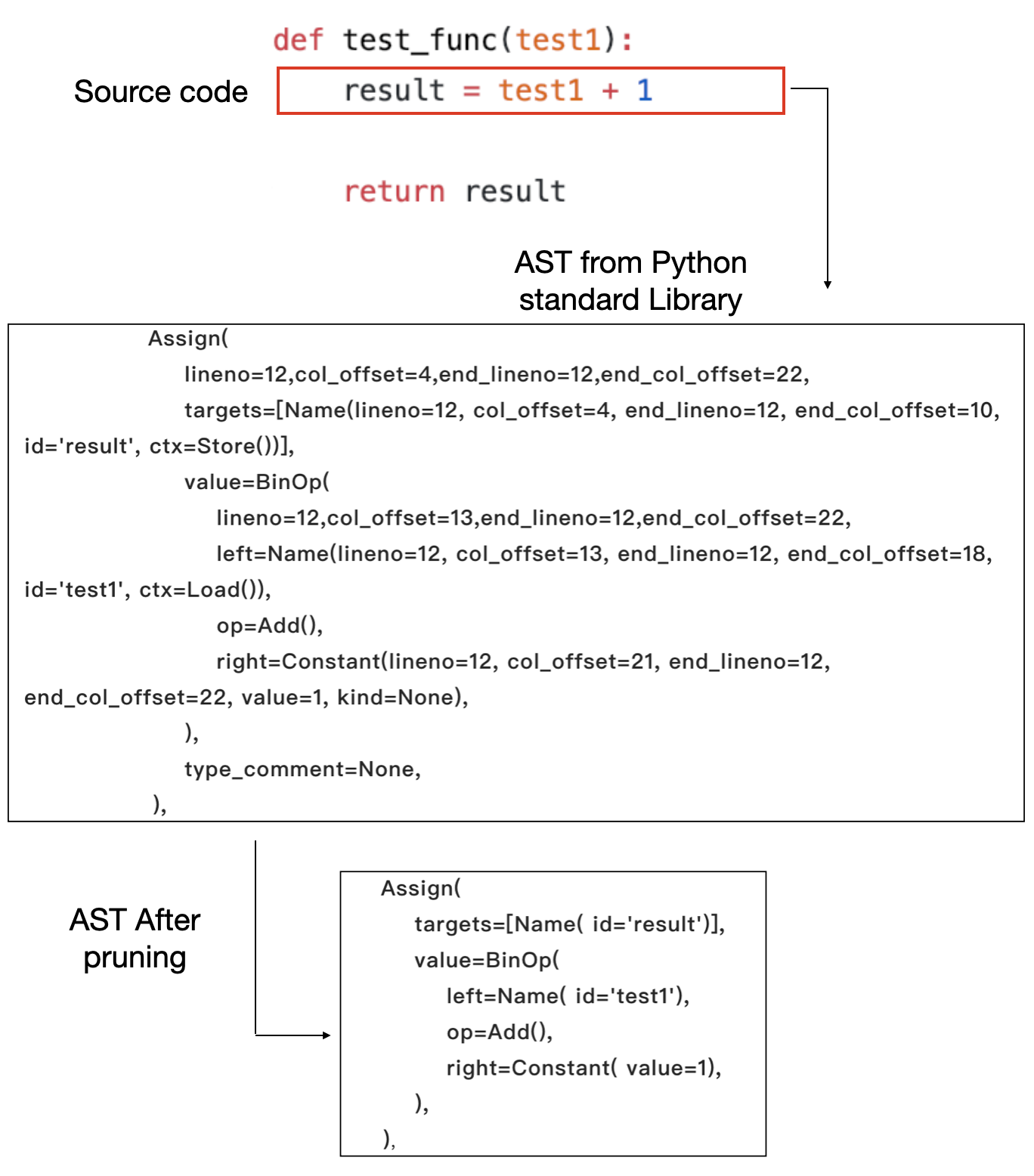}
		\caption{AST pruning process}
		\vspace{-3mm}
\end{figure}

In general, the length of AST from the compiled codes is greater than the plain source code, as shown in Figure 3, the AST from Python standard library contains a number of nodes such as \textit{lineno}, \textit{endlineno} and so on. 
% The AST for only very simple python code is very longer, which contains line number and some nodes in AST with \textit{None} information. 
Taking the snippet \textit{result = test1 + 1} as an example, both the original and pruned AST trees can be seen in Figure 3. It is clearly noticed that there exists a large amount of redundant information such as line number and code column offset in the original AST tree, leading to intractable AST exploration problem for large code corpus~\cite{wan2019multi}. Therefore, after generating AST, we will prune this tree by removing the meaningless and uninformed node to avoid unintended input for the model.

\subsection{AST Embedding Layer}
\begin{figure*}
\centering
        \includegraphics[scale=0.44]{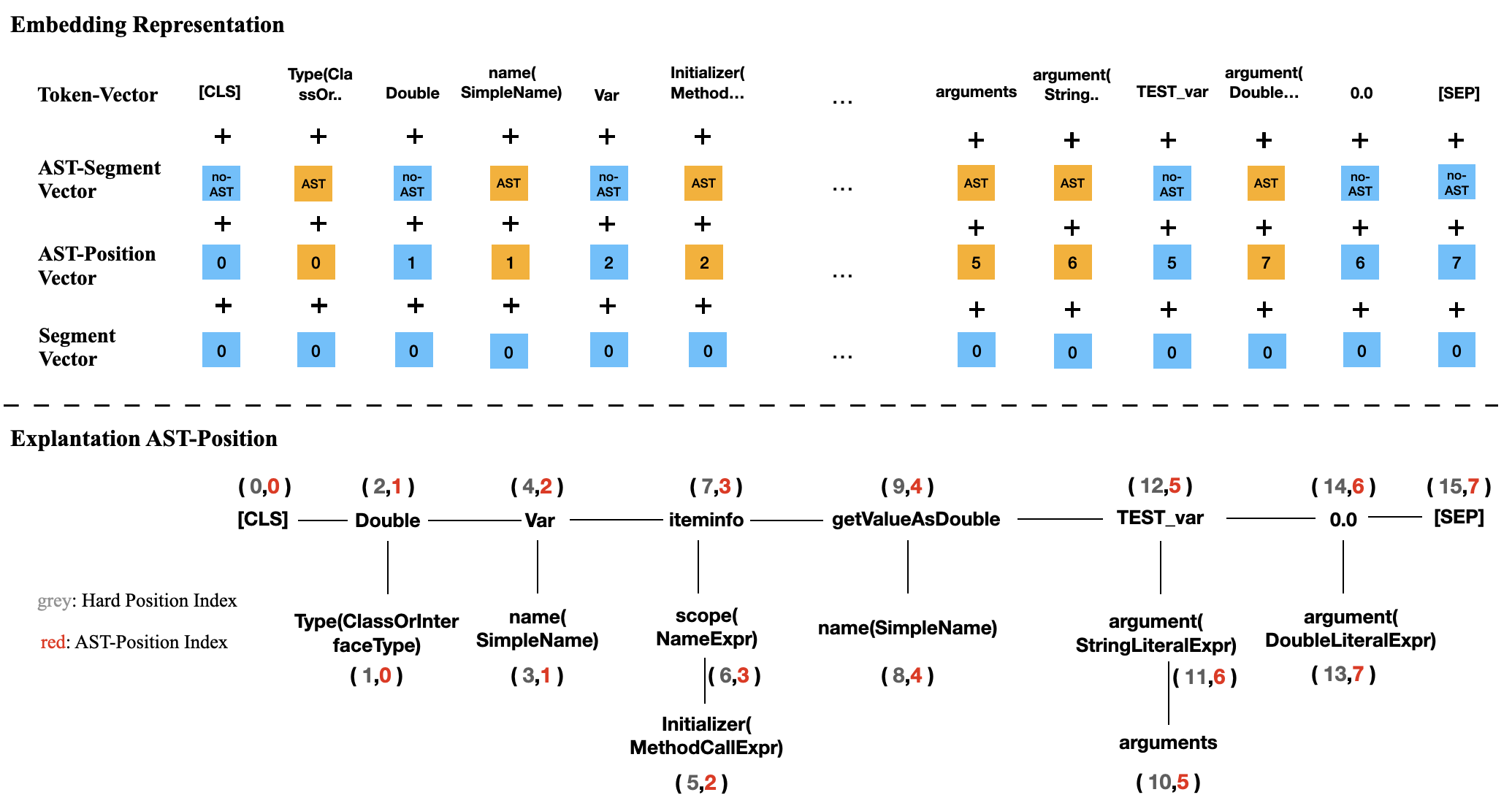}
		\caption{The overview of AST embedding representations}	
\end{figure*}

As mentioned above, we use the pruned AST as the input for model, and it will pass AST embedding layer first. The details of the AST embedding layer are unveiled in Figure 4, from which token-embedding vectors, AST-segment embedding vectors, AST-position embedding vectors and segment embedding vectors are generated. Taking the code snippet in Figure 2 as an example, we can see the additional AST information account for most of the tokens in the input, which unexpectedly causes changes in the meaning of the original code. To prevent this from happening, we use AST-segment embedding to distinguish between AST tokens and source code tokens. It is known that in BERT all the order information for input sequence is contained in the position embedding, allowing us to add different position information for input. 
% Taking the input in Figure 2 as an example, it is an AST for Java code described in subsection 3.1, we can see the additional AST information account for most of tokens in the input, which can lead changes in the meaning of the original code. To prevent this from happening, we use AST-Segment to 
Here, except for the AST-segment, we use an index combination of hard-position and AST-position to convey the order information. As seen in Figure 4, the index combination of \textit{name(SimpleName)} is \textit{(3,1)}, which means it locates at the 3rd position in the input sequence dimension while being the 1st AST token. In the front of \textit{name(SimpleName)}, there is only one extra AST token named \textit{Type(ClassOrInterfaceType)}. Segment embedding is similar to BERT. The output of the embedding layer is simply the sum of all embedding vectors from these four parts. The result is then passed into the AST-Mask transformer encoder to generate hidden vectors.

\subsection{AST-Mask Transformer}

\begin{figure}[t!]
\centering
        \includegraphics[scale=0.33]{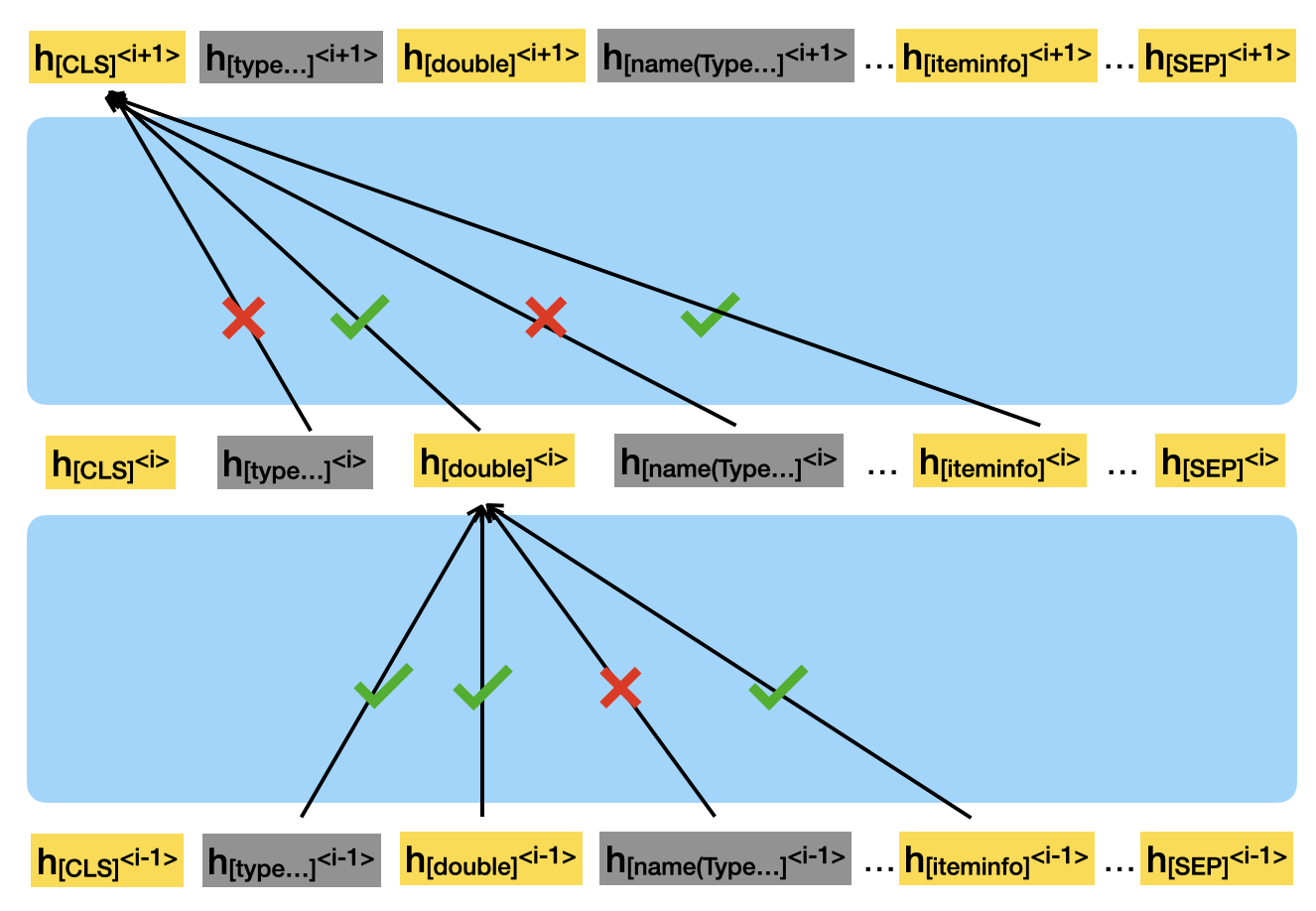}
		\caption{The explanation of AST-Mask-Transformer}
		\vspace{-6mm}
\end{figure}

Since the branch in AST contains the specific semantic knowledge to describe the role of the code token, it is rational to make AST tokens only contribute to the code tokens on the same branch. 
% the risk raised with AST information is that it can lead to changes in the meaning of the original sentence. 
For example, in Figure 2, [\textit{Type=ClassOrInterfaceType}] only describe the role of the [\textit{Double}] and has nothing to do with [\textit{Var}]. Therefore, the embedding of [\textit{Var}] should not be affected by [\textit{Type=ClassOrInterfaceType}]. As demonstrated in Figure 5, the \textit{Type=ClassOrInterfaceType} should not make a contribution to the embedding of [\textit{CLS}] tag that often used for classification bypass the [\textit{Double}]. This is because that the [\textit{Type=ClassOrInterfaceType}] is a tag in the branch of [\textit{Double}] and should only correlate to [\textit{Double}]. To prevent the AST information injection from changing the semantic of the input, AstBERT employs Mask-Self-Attention\cite{xu2021enabling} to limit the self-attention region in Transformer\cite{vaswani2017attention}. We use AST matrix \textit{M} to describe whether the AST token and code token are on the same branch, $M_{AST}$ is defined as follow:

\begin{equation}
    \centering
M_{AST_{i,j}}=\begin{cases}
1 & w_i \oplus w_j \\
0 & w_i \otimes w_j
\end{cases}
\end{equation}
where, $w_i \oplus w_j$ indicates that $w_i$ and $w_j$ are on the same AST branch, while $w_i \otimes w_j$ are not. $i$ and $j$ are the AST-position index. The AST mask matrix is then used to calculate the self-attention scores. Formally, the AST-mask-self-attention is defined as follow:

\begin{equation}
    Q^{i+1},K^{i+1},V^{i+1}=h^{i}W_{q},h^{i}W_{k},h^{i}W_v
\end{equation}

\begin{equation}
    S^{i+1}=softmax(\frac{K^{i+1^{T}}Q^{i+1}M_{AST}}{\sqrt{d_k}})
\end{equation}

\begin{equation}
    h^{i+1} = S^{i+1}V^{i+1}
\end{equation}
where $W_q$, $W_k$, $W_v$ are trainable model parameters. $h^i$ is the hidden state from the $i$th AST-mask-self-attention blocks. $d_k$ is the scaling factor. If $h^i_k$ and $h^i_j$ are not in same AST branch, the $M_{AST_{kj}}$ will make the attention score $S^{i+1}_{kj}$ to 0, which means $h^i_k$ makes no contribution to the hidden state of $h^i_j$.

% \centering
% $M_{i,j}=\begin{cases}
% 0 & w_i \oplus w_j \\
% -& w_i \otimes w_j
% \end{cases}$

% As shown in Figure 5, after this sequence encoded by transformer encoder, we use AST-Selection layer to mask the hidden vector from tokens of AST information and retain the hidden vector from tokens of original source code. This is because, even after pruning, among the all tokens in input, the tokens from AST information account for most of them. In the output layer, if we use all hidden vectors to finetune the downstream task, it will weaken the meaning of source code tokens. On other hand, transformer encoder has already fused the AST information for each source code token by attention mechanism. So here, we extract the hidden vectors of source code tokens by using AST-Selection layer. In output layer, if the task is classification, the hidden vectors after AST-Selection layer will pass average pooling layer to predict class label; If the task is named entity reorganization, each hidden vector after AST-Selection layer will be used predict tag directly.

We collect massive Python and Java codes from Alipay code repositories and generate the AST for these source code (Python code using standard AST API, Java code using Javaparser). We use these processed AST information to continue the pre-train of the model. The technique of pre-training is inspired by the masked language modeling (MLM), which is proposed by \cite{devlin2018bert} and proven effective.

\begin{table*}[]
\centering
\resizebox{0.98\linewidth}{!}{
\begin{tabular}{|ll|cccccc|}
\hline
\multicolumn{2}{|l|}{\multirow{2}{*}{}}                                & \multicolumn{6}{c|}{Datasets}                                                                                                                                                                                                              \\ \cline{3-8} 
\multicolumn{2}{|l|}{}                                                 & \multicolumn{5}{c|}{ACC}                                                                                                                                                                              & \multicolumn{1}{c|}{F1}            \\ \hline
\multicolumn{1}{|c|}{TASK}                             & Model         & \multicolumn{1}{l|}{AliCoQA}          & \multicolumn{1}{l|}{CoSQA}            & \multicolumn{1}{l|}{BFP\_{small}}              & \multicolumn{1}{l|}{BFP\_{medium}}              & \multicolumn{1}{l|}{AliCoRF}          & \multicolumn{1}{l|}{BigCloneBench} \\ \hline
\multicolumn{1}{|c|}{\multirow{4}{*}{Question Answering}}              & BERT          & \multicolumn{1}{c|}{0.402}            & \multicolumn{1}{c|}{0.399}            & \multicolumn{1}{c|}{\textbackslash{}} & \multicolumn{1}{c|}{\textbackslash{}} & \multicolumn{1}{c|}{\textbackslash{}} & \textbackslash{}                   \\ \cline{2-8} 
\multicolumn{1}{|c|}{}                                 & RoBERTA       & \multicolumn{1}{c|}{0.434}            & \multicolumn{1}{c|}{0.421}            & \multicolumn{1}{c|}{\textbackslash{}} & \multicolumn{1}{c|}{\textbackslash{}} & \multicolumn{1}{c|}{\textbackslash{}} & \textbackslash{}                   \\ \cline{2-8} 
\multicolumn{1}{|c|}{}                                 & CodeBERT      & \multicolumn{1}{c|}{0.532}            & \multicolumn{1}{c|}{0.526}            & \multicolumn{1}{c|}{\textbackslash{}} & \multicolumn{1}{c|}{\textbackslash{}} & \multicolumn{1}{c|}{\textbackslash{}} & \textbackslash{}                   \\ \cline{2-8} 
\multicolumn{1}{|c|}{}                                 & AstBERT       & \multicolumn{1}{c|}{\textbf{0.588}}   & \multicolumn{1}{c|}{\textbf{0.571}}   & \multicolumn{1}{c|}{\textbackslash{}} & \multicolumn{1}{c|}{\textbackslash{}} & \multicolumn{1}{c|}{\textbackslash{}} & \textbackslash{}                   \\ \hline
\multicolumn{1}{|c|}{\multirow{5}{*}{Code Refinement}} & LSTM          & \multicolumn{1}{c|}{\textbackslash{}} & \multicolumn{1}{c|}{\textbackslash{}} & \multicolumn{1}{c|}{0.100}            & \multicolumn{1}{c|}{0.025}            & \multicolumn{1}{c|}{0.111}            & \textbackslash{}                   \\ \cline{2-8} 
\multicolumn{1}{|l|}{}                                 & Transformer   & \multicolumn{1}{c|}{\textbackslash{}} & \multicolumn{1}{c|}{\textbackslash{}} & \multicolumn{1}{c|}{0.147}            & \multicolumn{1}{c|}{0.037}            & \multicolumn{1}{c|}{0.152}            & \textbackslash{}                   \\ \cline{2-8} 
\multicolumn{1}{|l|}{}                                 & CodeBERT      & \multicolumn{1}{c|}{\textbackslash{}} & \multicolumn{1}{c|}{\textbackslash{}} & \multicolumn{1}{c|}{0.164}            & \multicolumn{1}{c|}{0.052}            & \multicolumn{1}{c|}{0.176}            & \textbackslash{}                   \\ \cline{2-8} 
\multicolumn{1}{|l|}{}                                 & GraphCodeBERT & \multicolumn{1}{c|}{\textbackslash{}} & \multicolumn{1}{c|}{\textbackslash{}} & \multicolumn{1}{c|}{0.173}            & \multicolumn{1}{c|}{\textbf{0.091}}   & \multicolumn{1}{c|}{0.182}            & \textbackslash{}                   \\ \cline{2-8} 
\multicolumn{1}{|l|}{}                                 & AstBERT       & \multicolumn{1}{c|}{\textbackslash{}} & \multicolumn{1}{c|}{\textbackslash{}} & \multicolumn{1}{c|}{\textbf{0.176}}   & \multicolumn{1}{c|}{0.089}            & \multicolumn{1}{c|}{\textbf{0.183}}   & \textbackslash{}                   \\ \hline
\multicolumn{1}{|c|}{\multirow{7}{*}{Code Clone}}      & CDLH          & \multicolumn{1}{c|}{\textbackslash{}} & \multicolumn{1}{c|}{\textbackslash{}} & \multicolumn{1}{c|}{\textbackslash{}} & \multicolumn{1}{c|}{\textbackslash{}} & \multicolumn{1}{c|}{\textbackslash{}} & 0.820                              \\ \cline{2-8} 
\multicolumn{1}{|l|}{}                                 & ASTNN         & \multicolumn{1}{c|}{\textbackslash{}} & \multicolumn{1}{c|}{\textbackslash{}} & \multicolumn{1}{c|}{\textbackslash{}} & \multicolumn{1}{c|}{\textbackslash{}} & \multicolumn{1}{c|}{\textbackslash{}} & 0.930                              \\ \cline{2-8} 
\multicolumn{1}{|l|}{}                                 & FA-AST-GMN    & \multicolumn{1}{c|}{\textbackslash{}} & \multicolumn{1}{c|}{\textbackslash{}} & \multicolumn{1}{c|}{\textbackslash{}} & \multicolumn{1}{c|}{\textbackslash{}} & \multicolumn{1}{c|}{\textbackslash{}} & 0.950                              \\ \cline{2-8} 
\multicolumn{1}{|l|}{}                                 & RoBERTa       & \multicolumn{1}{c|}{\textbackslash{}} & \multicolumn{1}{c|}{\textbackslash{}} & \multicolumn{1}{c|}{\textbackslash{}} & \multicolumn{1}{c|}{\textbackslash{}} & \multicolumn{1}{c|}{\textbackslash{}} & 0.957                              \\ \cline{2-8} 
\multicolumn{1}{|l|}{}                                 & CodeBERT      & \multicolumn{1}{c|}{\textbackslash{}} & \multicolumn{1}{c|}{\textbackslash{}} & \multicolumn{1}{c|}{\textbackslash{}} & \multicolumn{1}{c|}{\textbackslash{}} & \multicolumn{1}{c|}{\textbackslash{}} & 0.965                              \\ \cline{2-8} 
\multicolumn{1}{|l|}{}                                 & GraphCodeBERT & \multicolumn{1}{c|}{\textbackslash{}} & \multicolumn{1}{c|}{\textbackslash{}} & \multicolumn{1}{c|}{\textbackslash{}} & \multicolumn{1}{c|}{\textbackslash{}} & \multicolumn{1}{c|}{\textbackslash{}} & 0.971                              \\ \cline{2-8} 
\multicolumn{1}{|l|}{}                                 & AstBERT       & \multicolumn{1}{c|}{\textbackslash{}} & \multicolumn{1}{c|}{\textbackslash{}} & \multicolumn{1}{c|}{\textbackslash{}} & \multicolumn{1}{c|}{\textbackslash{}} & \multicolumn{1}{c|}{\textbackslash{}} & \textbf{0.973}                     \\ \hline
\end{tabular}
}
\caption{Experiment results of different tasks on different dataset}
\label{tab:booktabscs}
\end{table*}

\section{Experiments}
We test the performance of our proposed model on different code understating tasks using the different released test datasets. We also look into the ablation studies.

\subsection{Dataset}

\textbf{Code Question Answering} CoSQA~\cite{huang2021cosqa} consists of 20,604 query-code pairs collected from the Microsoft Bing search engine. We randomly split CoSQA into 20,000 training and 604 validation examples. We also build AliCoQA dataset based on the code collected from the Alipay code repositories. We use the search logs from AntCode search engine as the source of queries and manually design heuristic rules to find the queries of code searching intent. For example, queries with the word of \textit{tutorial} or \textit{example} are likely to locate a programming description rather than a code function, so we remove such queries. Then, we use the CodeBERT matching model~\cite{feng2020codebert} to retrieve high-confidence codes for every query and manually check 5,000 query-code pairs to construct AliCoQA. We randomly split AliCoQA into 4,500 training and 500 validation samples.

\textbf{Code Clone Detection} We use BigCloneBench dataset~\cite{svajlenko2014towards} and discard samples with no labels. Finally, we randomly split it into 901,724 training set and 416,328 validation set.

\textbf{Code Refinement} BFP \cite{tufano2019empirical} dataset constains two subsets based on the code length. For BFP\_{small} dataset, the numbers of training and validation are 46,680 and 5,835, respectively. For the BFP\_{medium} dataset, the numbers of training and validation are 52,364 and 6,545. We collect code from Alipay code repositories and build AliCoRF dataset. Firstly, we identify commits having a message containing the words, such as \textit{fix}, \textit{solve}, \textit{bug}, \textit{problem} and \textit{issue}. Following that, for each bug-fixing commit, we extract the source code before and after the bug-fix. Finally, we manually check 9,000 bug-fix pairs to construct AliCoRF and randomly split it into 8,000 training set and 1,000 validation set.

\textbf{Evaluation Metric}
Following the settings in the previous work, we use accuracy as the evaluation metric on code question answering, and F1 score on code clone detection. We also use accuracy as the evaluation metric on code refinement, in which only the example being detected and fixed properly will be considered successfully completing the task. We give one example case for this task in Figure 6. In this example, the model successfully fixes the method name from \textit{getMin} to \textit{getMax}.

\begin{figure}
\centering
        \includegraphics[width=1.0\linewidth]{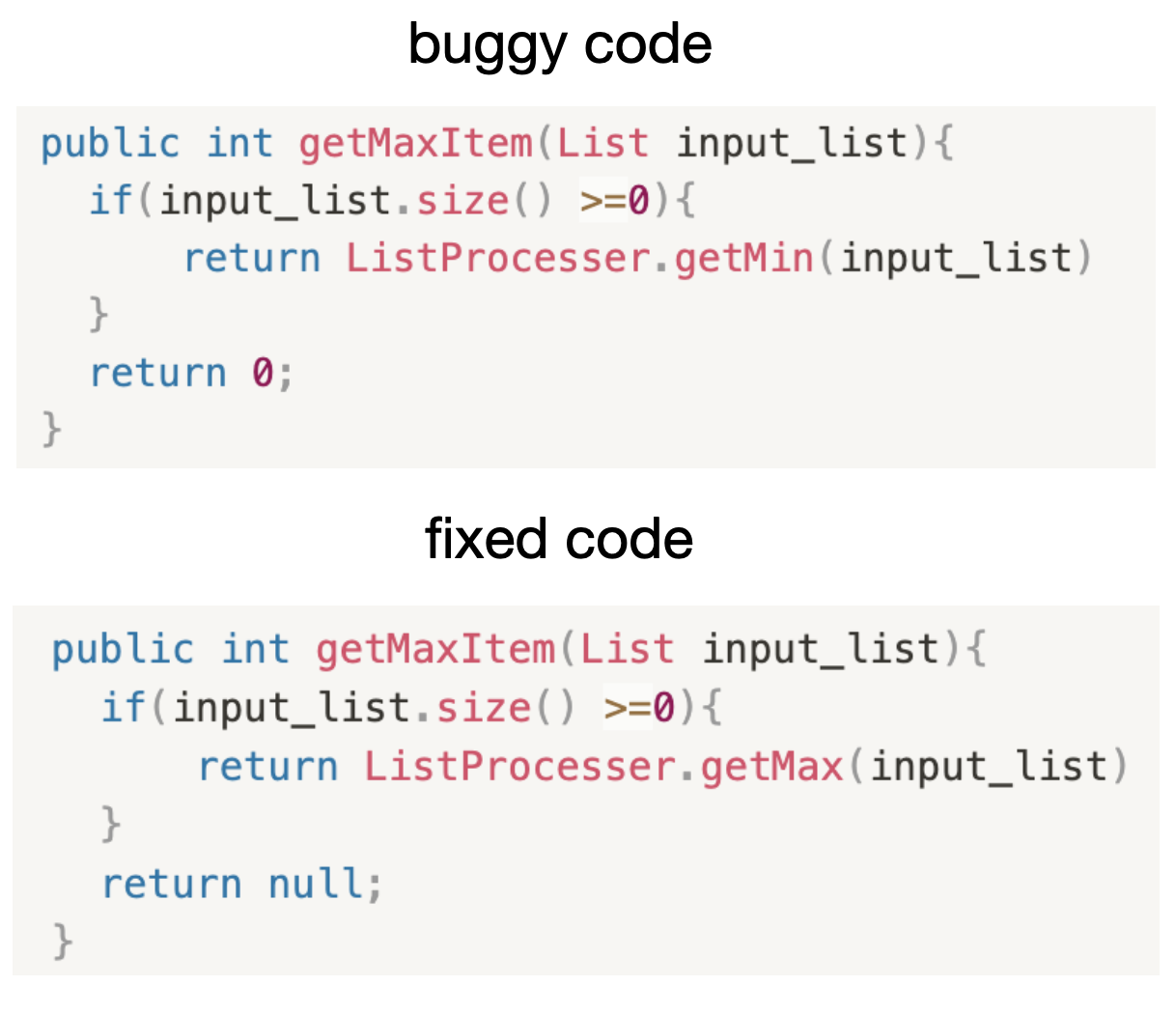}
		\caption{One case of AstBERT output for code refinement task.}	
\end{figure}

\subsection{Parameter Settings}
We follow the similar parameter settings in previous works~\cite{huang2021cosqa,svajlenko2014towards,tufano2019empirical}. On code question answering task, we set dropout rate to 0.1, maximum sequence length to 512, learning rate to 1e-5, warm-up rate to 0.1 and batch size to 16. On code clone detection task, learning rate is set to be  2e-5, batch size to be 16 and maximum sequence length to be 512. On code refinement task, we set learning rate to 1e-4, batch size to 32 and maximum sequence length to 256. For all experiments, we use the Adam optimizer to update model parameters~\cite{kingma2014adam}.

\subsection{Results of Code Question Answering }

% \begin{table}
% \centering
% \begin{tabular}{lcc}
% \toprule
% Model   &  Data         &Code Question Answering \\
% \midrule
% BERT         & CoSQA     & 39.92        \\
% RoBERTA      & CoSQA     & 42.12        \\
% CodeBERT       & CoSQA    & 52.65       \\
% AstBERT         & CoSQA     & 57.12     \\
% \bottomrule
% \end{tabular}
% \caption{Evaluation of different pre-trained language models on code question answering.}
% \label{tab:booktabscs}
% \end{table}

% \begin{table}
% \centering
% \begin{tabular}{ccc}
% \toprule
% Model   &  AliCoQA         &CoSQA \\
% \midrule
% BERT         & 40.20     & 39.92      \\
% RoBERTA      & 43.40     & 42.12        \\
% CodeBERT       & 53.20    & 52.65       \\
% AstBERT         &  58.80     & 57.12     \\
% \bottomrule
% \end{tabular}
% \caption{Evaluation of different pre-trained language models on code question answering.}
% \label{tab:booktabscs}

% \end{table}

% In addition to code NER task, we also conduct another code search experiment using CoSQA dataset which is proposed by \cite{huang2021cosqa}. 
We use CoSQA (Junjie Huang et al. 2021) dataset to verify the code question answering task. In this task, the test sample is the query-code pair and labeled as either “1” or “0”, indicating whether the code can answer the query. These query-code pairs are collected from Microsoft Bing search engine and annotated by human. We train different benchmark models using our dataset and evaluate the performance of each on CoSQA for code question answering:(i) BERT proposed by \cite{devlin2018bert}; (ii) RoBERTA proposed by \cite{liu2019roberta}; (iii) CodeBERT proposed by \cite{feng2020codebert}; and (iv) AstBERT. From Table 1, we can see the BERT and RoBERTA achieve a similar yet relative low Acc score in this task. This is because these two models are pre-trained by natural language corpus and not integrated with any code-related domain knowledge. CodeBERT achieves a better performance than the RoBERTA, similar to the results published by~\cite{huang2021cosqa}. Our AstBERT achieves the best performance compared with all benchmarks. This clearly demonstrates that the integration of AST information into the model can further improve model's ability for understanding semantic and syntactic information in the codes. We also evaluate our AstBERT on AliCoQA and the results show that in financial domain dataset AstBERT also achieves the best performance.

% \textcolor{red}{We also build AliCoQA dataset from Alipay code repositories. We use the search logs from AntCode search engine as the source of queries and manually desigin heuristic rules to find the queries of code searching intent. For example, queries with the word of \textit{tutorial} or \textit{example} are likely to seek a programming knowledge rather than a code function, so we remove such queries. Then, we use a CodeBERT matching model \cite{feng2020codebert} to retrieve high-confidence codes for every query and manually check 2000 query-code pairs to construct AliCoQA. We evaluate our AstBERT on AliCoQA and the results show that in financial domain dataset AstBERT also achieves the best performance.}

% In their paper, they proposed method CodeBERT performed better than RoBERTa by 8.43\% (52.87\% v.s. 40.34\%), which is a huge improvement. However, by adding the AST information into our AstBERT, our model performed even better than CodeBERT, and achieved a F1 score at 57.12\%. This proved the effectiveness of use AST information for code tasks.

\begin{table}[]
\centering
\resizebox{0.99\linewidth}{!}{
\begin{tabular}{|l|ccccc|}
\hline
\multirow{2}{*}{}            & \multicolumn{5}{c|}{Datasets}                                                                                                                \\ \cline{2-6} 
                             & \multicolumn{4}{c|}{ACC}                                                                                                     & F1            \\ \hline
Model                        & \multicolumn{1}{c|}{CoSQA} & \multicolumn{1}{c|}{AliCoQA}        & \multicolumn{1}{c|}{BFP}   & \multicolumn{1}{c|}{AliCoRF} & BigCloneBench \\ \hline
AstBERT                      & \multicolumn{1}{c|}{\textbf{0.571}} & \multicolumn{1}{c|}{\textbf{0.588}} & \multicolumn{1}{c|}{\textbf{0.176}} & \multicolumn{1}{c|}{\textbf{0.183}}   & \textbf{0.973}         \\ \hline
-w/o AST-position            & \multicolumn{1}{c|}{0.552} & \multicolumn{1}{c|}{0.558}          & \multicolumn{1}{c|}{0.174} & \multicolumn{1}{c|}{0.181}   & 0.970         \\ \hline
-w/o AST-Mask-Self-Attention & \multicolumn{1}{c|}{0.539} & \multicolumn{1}{c|}{0.544}          & \multicolumn{1}{c|}{0.165} & \multicolumn{1}{c|}{0.178}   & 0.966         \\ \hline
\end{tabular}
}
\caption{Ablation study}
\label{tab:booktabscs}
\vspace{-3mm}
\end{table}

\subsection{Results of Code Clone Detection }

Code clone detection is an another task when it comes to measuring the similarity of code-code pair, which can help reduce the cost of software maintenance. We use BigCloneBench  \cite{svajlenko2014towards} dataset for this task and treat this task as a binary classification to fine-tune AstBert. The experimental results are also shown in the Table 2. The \textbf{CDLH} model is proposed by \cite{wei2017supervised} to learn representations of code by AST-based LSTM and use hamming distance as optimization objective. The \textbf{ASTNN} model \cite{zhang2019novel} encodes AST subtrees by RNNs to learn representation for code. The \textbf{FA-AST-GMN} model \cite{wang2020detecting} uses a flow-augmented AST as the input and leverages GNNs to learn the representation for a program. The \textbf{GraphCodeBERT} \cite{guo2020graphcodebert}, which is a pre-trained model using data flow at the pre-training stage to leverage the semantic-level structure of code, learns the representation of code. The experiment shows that our AstBERT achieves the best results in code clone detection task.

\subsection{Results of Code Refinement}

% \begin{table}
% \centering
% \resizebox{\linewidth}{!}{
% \begin{tabular}{lccc}
% \toprule
% Model      & BFP\_small    &BFP\_medium   &AliCoRF \\
% \midrule
% LSTM            & 10        &  2.5     & 11.1        \\
% Transformer          & 14.7      & 3.7    & 15.2     \\
% CodeBERT         & 16.4         & 5.2   & 17.6        \\
% GraphCodeBERT              & 17.3    & 9.1   & 18.2 \\
% AstBERT              & 17.6       &8.9        & 18.3 \\
% \bottomrule
% \end{tabular}}

% \caption{Evaluation of different models on code refinement}
% \label{tab:booktabscs}
% \end{table}

% \begin{table*}
% \centering
% \resizebox{\linewidth}{!}{
% \begin{tabular}{lcccccccccc}
% \toprule
% Methods                      & Data    & Acc        & Data          & Acc     & Data     & F1    & Data          & Acc       & Data  & Acc\\
% \midrule
% AstBERT                      & CoSQA   &  57.12     & AliCoQA      & 58.8     & BigCloneBench & 0.973  & BFP_{small}   & 17.6   &AliCoRF  & 18.3 \\
% -w/o AST-position            & CoSQA   & 55.26      & AliCoQA      & 55.8      & BigCloneBench & 0.970 & BFP_{small}    & 17.4    &AliCoRF  & 18.1\\
% -w/o AST-Mask-Self-Attention & CoSQA    & 53.78     & AliCoQA      & 54.4      & BigCloneBench  & 0.966 & BFP_{small}    & 16.5   &AliCoRF   & 17.8 \\
% \bottomrule
% \end{tabular}
% }
% \caption{Ablation study}
% \label{tab:booktabscs}
% \end{table*}

In general, code refinement is the task of locating code defects and automatically fixing them, which has been considered critical to uncovering any financial risks. We use both BFP\_{small} and BFP\_{medium} datasets \cite{tufano2019empirical} to verify the performance of all models and show results in the Table 1. This is a Seq2Seq task, and we record relevant accuracy for each benchmark model. We take the results of \textbf{LSTM} and \textbf{Transformer} as recorded in \cite{guo2020graphcodebert}. It is observed in the table that \textbf{Transformer} outperforms \textbf{LSTM}, which indicates that \textbf{Transformer} has a better ability of learning the representation of code. Both \textbf{CodeBERT} and \textbf{GraphCodeBERT} are pre-trained models, which present state-of-the-art results at their time. Our \textbf{AstBERT} achieves a better performance than other pre-trained models on BFP\_{small} dataset, while obtaining the competitive result on BFP\_{medium} dataset. This again demonstrates the effectiveness of incorporating the AST information in the pre-trained model is helpful to the code understanding, including the code refinement task.

\subsection{Ablation Studies}

In this subsection, we explore the effects of the AST-position and AST-Mask-Self-Attention for AstBERT on three tasks. ``\textbf{w/o AST-position}'' refers to fine-tuning AstBERT without AST-position. ``\textbf{w/o AST-Mask-Self-attention}'' means that each token in input, regardless of its position in the AST tree, calculates the attention scores with other tokens. As shown in Table 2, we have made the following observations: (i) Without AST-position or AST-Mask-Self-Attention, the performance of AstBERT on code question answering has shown a clear decline; (ii) It also can be seen that the model without AST-Mask-Self-Attention demonstrates an even worse performance than without AST-position, which confirms sufficient AST tokens can help incorporate the syntactic structures of the code. The same trend can also be observed on code clone detection and code refinement. We can conclude that the AST-position and the AST-Mask-Self-Attention play a pivotal role in incorporating the AST information into the model.

\section{Conclusion}
In this paper, we propose AstBERT, a simple and effective way to enable pre-trained language model for financial code understanding by integrating semantic information from the abstract syntax tree (AST). In order to encode the structural information, AstBERT uses a designated AST-Segment and AST-Position in the embedding layer to make model incorporate such AST information. Following that, we propose the AST-Mask-Self-Attention to limit the region when calculating attention scores, preventing the input from deviating from its original meaning. 
% We also develop a new dataset CoNER for code named entity recognition(NER) task, which to the best of our knowledge is the first dataset for code NER task. 
We conduct three different code understanding related tasks to evaluate the performance of the AstBERT. The experiment results show that AstBERT outperforms baseline models on both code question answering and clone detection. For code refinement task, the model achieves state-of-the-art performance on BFP\_{small} dataset and competitive performance on BFP\_{medium} dataset.

\bibliography{emnlp2022}
\bibliographystyle{acl_natbib}

% \appendix

% \section{Example Appendix}
% \label{sec:appendix}

% This is a section in the appendix.

\end{document}